\documentclass{article} %
\usepackage{colm2024_conference}

\usepackage{microtype}
\usepackage{linguex}
\usepackage{hyperref}
\usepackage{url}
\usepackage{booktabs}
\usepackage{pifont}
\usepackage{multirow}
\usepackage{colortbl}
\usepackage{graphicx}
\usepackage{csquotes}
\usepackage[T1]{fontenc}
\usepackage{upquote}

\newcommand{\cmark}{\ding{51}}%
\newcommand{\xmark}{\ding{55}}

\definecolor{darkblue}{rgb}{0, 0, 0.5}
\hypersetup{colorlinks=true, citecolor=darkblue, linkcolor=darkblue, urlcolor=darkblue}

\title{Code Pretraining Improves Entity Tracking Abilities of\\ Language Models}

\author{Najoung Kim* \\
Department of Linguistics\\
Boston University\\
\texttt{najoung@bu.edu} \\
\And
Sebastian Schuster* \\
Department of Linguistics \\
University College London \\
\texttt{s.schuster@ucl.ac.uk} \\
\And
Shubham Toshniwal* \\
NVIDIA \\
\texttt{stoshniwal@nvidia.com}
}

\colmfinalcopy %
\begin{document}

\maketitle

\begin{abstract}
Recent work has provided indirect evidence that pretraining language models on code improves the ability of models to track state changes of discourse entities expressed in natural language. In this work, we systematically test this claim by comparing pairs of language models on their entity tracking performance. Critically, the pairs consist of base models and models trained on top of these base models with additional code data. We extend this analysis to additionally examine the effect of math training, another highly structured data type, and alignment tuning, an important step for enhancing the usability of models. We find clear evidence that models additionally trained on large amounts of code outperform the base models. On the other hand, we find no consistent benefit of additional math training or alignment tuning across various model families.
\end{abstract}

{\let\thefootnote\relax\footnotetext{\hspace{-0.2cm}$^*$Equal contribution.}}

\section{Introduction}

Entity tracking, the capacity to track how properties of discourse entities and their relationships change as a discourse unfolds, is an important ability for understanding longer contexts as well as other critical capabilities such as planning. For example, to successfully parse the following recipe, an agent needs to track what happens to the different entities, such as ingredients.

\ex. \label{ex:recipe} Put the eggs, sugar, flour, and baking powder in a bowl and mix to form a light batter. Make sure that the final batter does not contain any lumps of flour or sugar.

\cite{kim-schuster-2023-entity} showed that several Transformer-based large language models (LLMs), such as GPT-3.5, exhibit a non-trivial entity tracking capacity. At the same time, they found that similar models, such as GPT-3, seem to lack this ability. Based on the limited information available about the differences between the GPT-3 and GPT-3.5 models, \cite{kim-schuster-2023-entity} hypothesized that pretraining on large amounts of code imbues LLMs with entity tracking abilities. However, due to the opacity of training data specifications of these models, it remains unclear whether code pretraining indeed is the critical difference. In this work, we re-evaluate this claim with open-source LLMs for which more information about the pretraining process is available. We extend our analysis to the effect of math and instruction tuning as well as code. Upon comparing pairs of base models and models additionally trained on code, math, or alignment tuned,\footnote{We use the term alignment tuning to refer to various methods of making language models more useful for interactive settings, including supervised instruction finetuning (SFT)~\citep{wei2022finetuned}, reinforcement learning from human feedback (RLHF)~\citep{ouyang2022training}, and direct preference optimization (DPO)~\citep{rafailov2023direct}.
} we find a clear benefit of code training but no consistent benefit of math training or alignment tuning.

\section{Related work}
Including code in the pretraining data mixture, even for models not explicitly specialized for code, has become increasingly customary in LLM training \citep[\textit{i.a.}]{chowdhery2023palm,touvron2023llama,geminiteam2024gemini,groeneveld2024olmo}. In addition to serving the popular use case of LLMs in code completion and generation \citep{chen2021evaluating}, adding code to the pretraining data mixture has been claimed to improve general reasoning capacities of LLMs \citep{fu2022gptroadmap,ma2023training,yang2024llm}. \citet{kim-schuster-2023-entity} hypothesized that a concrete capacity that can benefit from code is entity tracking: converging evidence towards this claim is contributed by observations from \citet{madaan2022language} (code pretrained models like Codex perform better than models primarily trained on language data on ProPara \citep{dalvi-etal-2019-everything}), \citet{sap-etal-2022-neural} (GPT-3.5 performs better on object tracking than GPT-3), and \citet{muennighoff2023scaling} (adding code to the pretraining data improves performance the on bAbI tasks~\citep{weston2016towards}). Furthermore, \citet{prakash2024finetuning} observed that a base model finetuned on arithmetic tasks improved performance on a simplified version of the entity tracking task by \citet{kim-schuster-2023-entity}, suggesting that structured data in general beyond code may contribute to the development of an entity tracking capacity in language models.

\begin{table}[t]
\resizebox{1\columnwidth}{!}{
    \centering
    \renewcommand{\arraystretch}{1.1}
    \begin{tabular}{l l c c c c l}
    \toprule
        \multirow{2}{*}{\shortstack[l]{\textbf{Exp}}} & \multirow{2}{*}{\shortstack[l]{\textbf{Model}}} & \multirow{2}{*}{\shortstack[l]{\textbf{Size}}} & \multicolumn{3}{c}{\textbf{Additional Training}} & \multirow{2}{*}{\shortstack[l]{\textbf{Base Model}}} \\
        \cmidrule{4-6}
         & & & +Code & +Math & +Instruct/Chat \\
        \midrule
         Exp 1  & Llama 2 & 7B--70B & & & & - \\
         (code)& Code Llama & 7B--70B & \ding{51} & & & Llama 2 \\   \arrayrulecolor{lightgray}\cline{2-7}
               & DeepSeek & 7B & & & & - \\ 
               & DeepSeek-Coder & 7B & \ding{51} & & & DeepSeek \\ \cline{2-7}
               & Gemma & 8B & & & & - \\ 
               & CodeGemma & 8B & \ding{51} & & & Gemma \\ 
               \arrayrulecolor{black}\midrule

         Exp 2  & Llama & 7B & & & & -\\
         (math) & FLoat & 7B & & \ding{51} (instruct) & & Llama \\ \arrayrulecolor{lightgray} \cline{2-7}
               & Mistral & 7B & & & & - \\
               & OpenMathMistral & 7B & & \ding{51} (instruct) & & Mistral \\ \cline{2-7}
               & DeepSeek-Coder& 7B & \ding{51} & & & DeepSeek \\
               & DeepSeek-Math & 7B & \ding{51} & \ding{51} & & DeepSeek-Coder \\ \cline{2-7}
               & Code Llama & 7B, 34B & \ding{51} & & & Llama 2 \\ 
               & Llemma & 7B, 34B & \ding{51} & \ding{51} & & Code Llama \\

               \arrayrulecolor{black}\midrule
               
                Exp 3  & Llama 2 & 7B--70B & & & & -\\
                (alignment) & Llama 2-Chat & 7B--70B & & & \ding{51} & Llama 2 \\ \arrayrulecolor{lightgray}\cline{2-7}
                & Code Llama & 7B--70B & \ding{51} & & & Llama 2  \\
                & Code Llama-Instruct & 7B--70B & \ding{51} & & \ding{51} & Code Llama 2 \\ \cline{2-7}
                & Gemma & 8B &  & & & -  \\
                & Gemma-Instruct & 8B &  & & \ding{51} & Gemma \\ \cline{2-7}                
                & CodeGemma & 8B & \ding{51} & & & Gemma  \\
                & CodeGemma-Instruct & 8B & \ding{51} & & \ding{51} & CodeGemma \\ \cline{2-7}
               & DeepSeek & 7B & & & & - \\ 
               & DeepSeek-Chat & 7B  & & & \ding{51} & DeepSeek \\ 
               \cline{2-7}
              & DeepSeek-Coder & 7B & \ding{51} & & & DeepSeek \\ 
               & DeepSeek-Coder-Instruct & 7B & \ding{51} & & \ding{51} & DeepSeek-Coder \\ 

               \arrayrulecolor{black}\bottomrule
    \end{tabular}
    }
    \caption{Summary of the models compared and their pretraining data composition.} 
    \label{tab:models}
\end{table}

\section{Experiments}
We aim to systematically test the hypothesis that code pretraining leads to better entity tracking put forward by \citet{kim-schuster-2023-entity}, through a series of experiments comparing base models and models continued to be trained on code on top of the base models. We additionally test the hypothesis that pretraining on math, another type of structured data, leads to better entity tracking performance through similar comparisons.

\subsection{Models}
We selected model pairs that have been reported to vary only in terms of their pretraining data. For testing the code hypothesis, we compared the following pairs of models: (Llama 2, Code Llama), (DeepSeek, DeepSeek-Coder), and (Gemma, CodeGemma), where the second model in each pair is obtained by continuing to train the first model on additional code data. We tested 7B, 13B, and 70B models in the Llama 2 series. For testing the math hypothesis, we compared the following four pairs of models: (Code Llama, Llemma), (DeepSeek-Coder, DeepSeek-Math), (Llama, FLoat), and (Mistral, OpenMathMistral). Again, the second model in each pair is obtained by training the first model on additional math data. For alignment tuning, we compared (Llama 2, Llama 2-chat), (Code Llama, Code Llama-Instruct), (Gemma, Gemma-Instruct), (CodeGemma, CodeGemma-Instruct), (DeepSeek, DeepSeek-Chat), and (DeepSeek-Coder, DeepSeek-Coder-Instruct). These comparisons are summarized in Table~\ref{tab:models}. See Appendix~\ref{app:model-details}, Table~\ref{tab:model_details} for more details about the models.

\begin{table}[t]
\begin{tabular}{p{0.9\linewidth}}
    \toprule
    \textbf{2-shot prompt} \\
    \midrule
    Given the description after "Description:", write a true statement about all boxes and their contents to the description after "Statement:".\\\\
    
    Description: Box 0 contains the car, Box 1 contains the cross, Box 2 contains the bag and the machine, Box 3 contains the paper and the string, Box 4 contains the bill, Box 5 contains the apple and the cash and the glass, Box 6 contains the bottle and the map.\\
    Statement: Box 0 contains the car, Box 1 contains the cross, Box 2 contains the bag and the machine, Box 3 contains the paper and the string, Box 4 contains the bill, Box 5 contains the apple and the cash and the glass, Box 6 contains the bottle and the map.\\\\
    
    Description: Box 0 contains the car, Box 1 contains the cross, Box 2 contains the bag and the machine, Box 3 contains the paper and the string, Box 4 contains the bill, Box 5 contains the apple and the cash and the glass, Box 6 contains the bottle and the map. Remove the car from Box 0. Remove the paper and the string from Box 3. Put the plane into Box 0. Move the map from Box 6 to Box 2. Remove the bill from Box 4. Put the coat into Box 3.\\
    Statement: Box 0 contains the plane, Box 1 contains the cross, Box 2 contains the bag and the machine and the map, Box 3 contains the coat, Box 4 contains nothing, Box 5 contains the apple and the cash and the glass, Box 6 contains the bottle.\\\\
    
    Description: \{\texttt{description}\}\\
    Statement: Box 0 contains\\
    \bottomrule
\end{tabular}
\caption{Prompts with 2-shot in-context demonstrations.}
\label{tab:in-context-prompts}
\end{table}

\subsection{Evaluation setup}

We adopted the boxes task (the ``base''
version) from \citet{kim-schuster-2023-entity} for testing the models' entity tracking capacity. In this task, the input to the LLM is a textual description of the contents of seven boxes followed by 1--12 descriptions of operations that change the contents of the individual boxes. In response to this input, the LLM is prompted to state the contents of each box according to the initial description and the state-changing operations. We used the same prompt and 2-shot in-context learning examples as \cite{kim-schuster-2023-entity} (see Table~\ref{tab:in-context-prompts} for an example). We used a slightly different prompt format for chat-optimized models to align the task better to the input format the models were trained on. The inputs to the model are provided as ``user'' prompts, and the expected model outputs are formatted as ``assistant'' (see  Table~\ref{tab:in-context-prompts-chat} in Appendix for an example prompt). 

We noticed that smaller models suffered from formatting issues, often deviating from the format specified by the prompt or omitting the contents of some boxes. For this reason, we used regular expression-based constrained decoding using the \texttt{outlines} library~\citep{Willard2023EfficientGG}.\footnote{\url{https://github.com/outlines-dev/outlines}}

We report all results divided into the number of operations affecting the target box rather than reporting one aggregate accuracy metric. This is to distinguish trivial cases from cases that actually require tracking state changes---when the number of operations affecting the target box is 0, simply copying from the initial state description yields the correct answer. Furthermore, we compare the model results to the strong random baseline by \cite{kim-schuster-2023-entity}. For this baseline, we randomly sample 0 to 3 objects for each box from the set of objects that have been previously mentioned in a clause with the box in question.

\begin{figure}
    \centering
    \includegraphics[width=\textwidth]{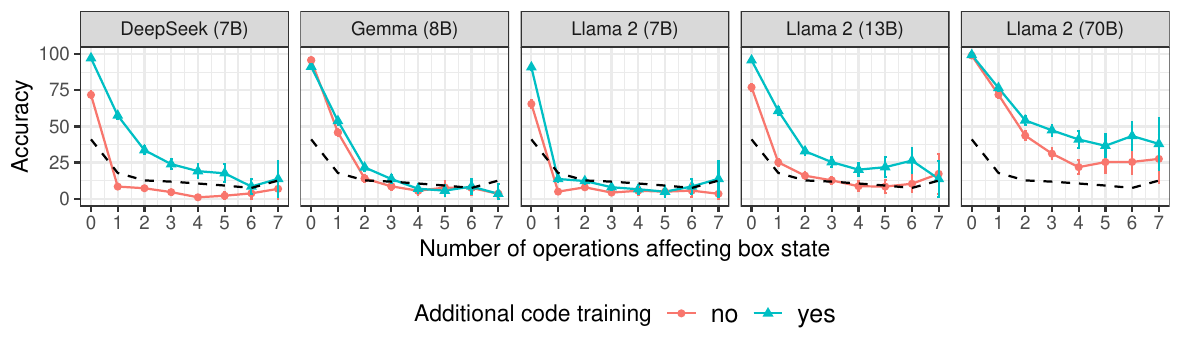}
    \caption{Entity tracking results for DeepSeek, Gemma, and Llama 2 models. Error bars indicate 95\% confidence intervals, and the black dashed lines show the performance of the random baseline.}
    \label{fig:base-vs-code}
\end{figure}

\subsection{Experiment 1: Effect of Code}

Figure~\ref{fig:base-vs-code} compares the entity tracking performance of base models (red lines) and code models (blue lines) for models from the DeepSeek  \citep{DeepSeekai2024DeepSeek,guo2024DeepSeekcoder}, Gemma \citep{gemmateam2024gemma}, and Llama 2 families of various sizes \citep{touvron2023llama, rozière2024code}. In general, we find clear evidence that continued training on large amounts of code improves entity tracking abilities, as can be seen for the Llama 2 13B and 70B models as well as for the DeepSeek models. In these model comparisons, the models trained on code consistently outperformed the base models on the nontrivial cases of entity tracking (number of operations affecting box state $\geq$ 1). In the case of 13B models, a boost in trivial cases is also observed (number of operations = 0); in 70B models, performance on the trivial cases is already saturated in the base model.

In Llama 2 7B models, the gains through additional code training are relatively minor, with most of the gains deriving from boosts in examples where the number of operations is either 0 or 1. Similarly minor gains were observed in CodeGemma 8B, except that the gains were observed in examples with 1 and 2 operations. For both of these models, we also observed that for number of operations greater than 0 (Llama 2 7B) and 2 (Gemma 8B), neither the base nor the code variants perform better than our random baseline. These results suggest that there is both a possible effect of scale in the effectiveness of code training as observed in the Llama 2 series, and an effect of the amount of additional code training (DeepSeek-Coder: 2T tokens, Code Llama: 500B tokens).

\subsection{Experiment 2: Effect of Math}

\begin{figure}
    \centering
    \includegraphics[width=\textwidth]{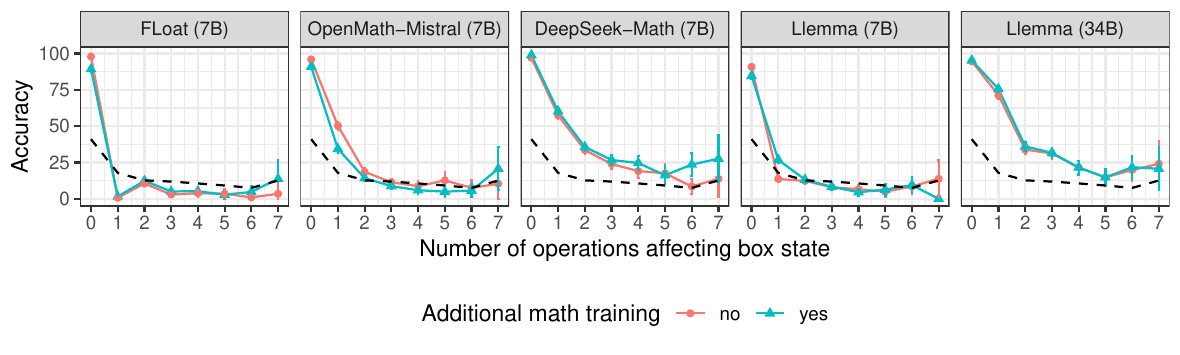}
    \caption{Entity tracking results for models trained with additional math data. See Table~\ref{tab:models} for the model names of the base and math models. Error bars indicate 95\% confidence intervals, and the black dashed lines show the performance of the random baseline.}
    \label{fig:math-tuning}
\end{figure}

In evaluating the effect of additional math training, we start by revisiting the claim of \citet{prakash2024finetuning} that the FLoat model obtained by finetuning Llama 7B \citep{touvron2023llama1} on arithmetic tasks from \citet{liu2023goat} yields superior entity tracking performance. As can be seen in  Table~\ref{tab:FLoat}, FLoat did show slightly higher accuracy on the non-trivial tracking cases (number of operations $\geq 1$) than the base model, but the gain was marginal.\footnote{These numbers are not expected to align with numbers reported in \citet{prakash2024finetuning} because they used a modified version of the original task.} Furthermore, neither the base Llama model nor the FLoat model performed better than our random baseline on non-trivial entity tracking examples (Figure~\ref{fig:math-tuning}, far left).

\begin{table}[h]
    \centering
    \renewcommand{\arraystretch}{1.1}
    \begin{tabular}{cccc}
    \toprule
        \textbf{Model} & \textbf{Aggregate} & \textbf{NumOps = 0} & \textbf{NumOps $\geq 1$} \\\midrule
        Llama 7B & 28.67 & 97.93 & 4.34 \\
        FLoat 7B & 27.55 & 89.33 & 5.85 \\\bottomrule
    \end{tabular}
    \caption{Llama 7B vs. FLoat 7B results.}
    \label{tab:FLoat}
\end{table}

Following this observation, we compared Mistral and OpenMathMistral models where the latter is a model further trained on OpenMathInstruct-1, a synthetically generated instruction-tuning dataset containing 1.8M unique solutions to math problems sourced from MATH and GSM8K datasets~\citep{toshniwal2024openmathinstruct1}. As shown in Figure~\ref{fig:math-tuning}, OpenMathMistral only achieved marginal gains over the base Mistral model when there are 7 operations affecting the target box, and in most other cases, the base model consistently outperforms the math-finetuned model. Furthermore, neither model outperformed the random baseline for examples with more than 2 operations affecting the box of interest. 

The unclear benefit of additional math training is further corroborated by marginal gains in models trained on math data that are not in ``instruct'' format like FLoat and OpenMathMistral. Figure~\ref{fig:math-tuning} shows that DeepSeek-Math \citep{shao2024deepseekmath} performed close to the DeepSeek-Coder model for most cases. The gains are even more limited in the comparison between Code Llama vs Llemma \citep{azerbayev2024llemma}. Llemma 34B outperformed Code Llama 34B by a narrow margin for the non-trivial tracking cases (Llemma: 47.86, Code Llama: 45.46 for examples where the number of operations $\geq 1$). These results suggest a limited benefit of additional math pretraining on entity tracking.

\subsection{Experiment 3: Effect of Alignment Tuning}

\begin{figure}
    \centering
    \includegraphics[width=\textwidth]{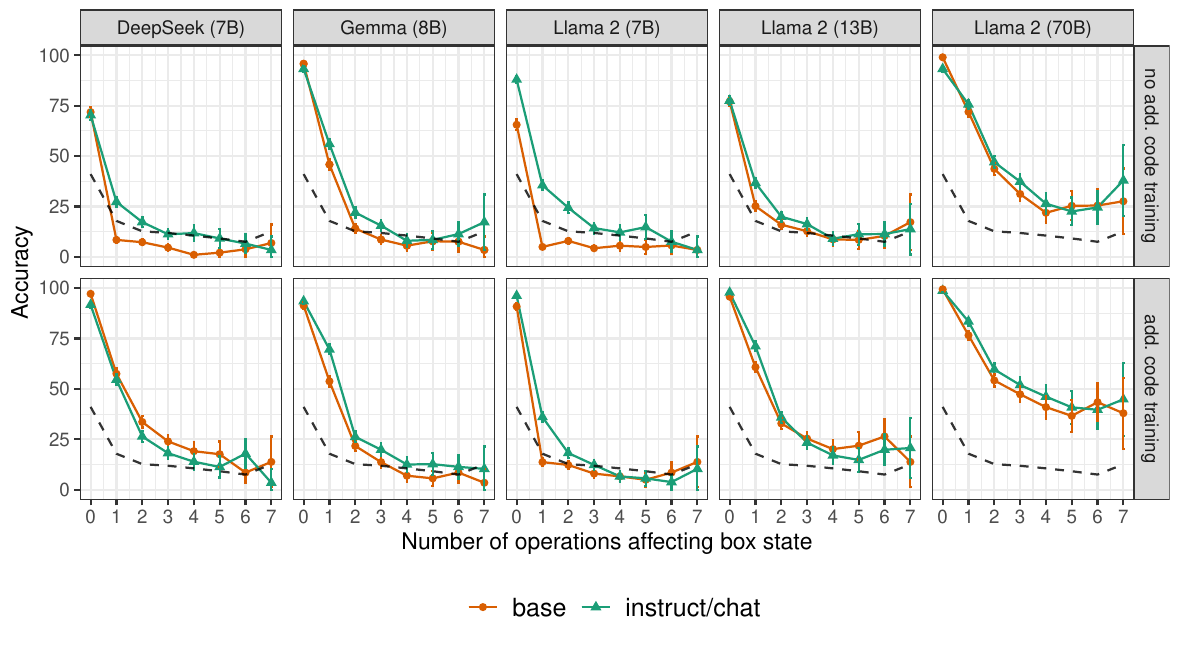}
    \caption{Entity tracking results for alignment-tuned DeepSeek, Gemma and Llama 2 models. The top panels show models without additional code training, whereas the bottom panels show models that have been trained on additional amounts of code before alignment tuning. Error bars indicate 95\% confidence intervals, and the black dashed lines show the performance of the random baseline.
    }
    \label{fig:instruction-tuning}
\end{figure}

Finally, we explore the effect of alignment tuning on entity tracking. For models in the Llama 2 family, alignment tuning the base models led to minor gains (Figure~\ref{fig:instruction-tuning}, orange  vs. green lines in the top row panels), whereas alignment-tuned code models did not consistently lead to gains and sometimes performed worse than the non-alignment-tuned counterparts (orange vs. green lines in the bottom row panels). Nevertheless, the best-performing model was CodeLlama 70B-instruct (64.9 accuracy on 1+ operations), combining code and alignment tuning. 

The DeepSeek models showed similar trends to the general observation made above: alignment tuning of the base model led to gains, whereas alignment tuning of the code model did not. Gemma 8B and CodeGemma 8B models did benefit from alignment tuning, similarly to Llama 2 7B and CodeLlama 7B models, although the gains were smaller.

Overall, alignment tuning affects base and code models differently, where the gains for base models tend to be greater. The benefit of alignment tuning for base models seems to be inversely correlated with scale: smaller base models benefit more from alignment tuning.

\section{Conclusion and Future Work}

We explored the effect of code, math, and alignment tuning on LLMs' capacity to track entities in natural language text. Our main findings are threefold:

\begin{enumerate}
    \item Additional code training leads to consistent improvements across model families and sizes.
    \item Additional math training does not yield consistent improvements, and the performance gains are at best marginal.
    \item Alignment tuning leads to different patterns of improvement depending on whether it was applied to base models or code models. Base models consistently benefit from alignment tuning, and smaller models see more improvement. The benefit for code models is more mixed, but the best performance is achieved through combining code and instruction tuning.
\end{enumerate}

Our work thus adds to a growing body of literature that suggests that pretraining on code improves LLM performance on reasoning tasks, including commonsense reasoning~\citep{madaan2022language}, chain-of-thought reasoning \citep{wei2022chain},
mathematical problems \citep{razeghi2024backtracking}, and entity tracking tasks \citep{muennighoff2023scaling}. \emph{Why might this be the case?} \cite{kim-schuster-2023-entity} argued that keeping track of the states of variables is important for producing correct code, and hypothesized that this kind of procedural input may provide a stronger training signal than pure natural language text. We consider investigating how code training imbues models with entity tracking and other reasoning abilities an important direction for future research.

\paragraph{Limitations} While the pairs of models we compared are ``minimal pairs'', several possible confounds remain in our interpretation. For example, we interpreted parts of the results in Experiment 2 as marginal benefit of math compared to code, but the OpenMathInstruct dataset (1.5 GB) is two orders of magnitude smaller in terms of the number of tokens compared to Code Llama's code data (500 B tokens), 
 so the size of the additional training data could be a confound. The additional math training data of DeepSeek-Math is more comparable (120B tokens), but we do not have a model that is only continually trained on math data; DeepSeek-Math is trained on both code and math. Furthermore, the math data vary along several other important dimensions: OpenMathInstruct, and FLoat models use synthetic data, whereas others use naturally occurring data. FLoat and OpenMathMistral are tuned on math data in instruction format, whereas the training data of DeepSeek-Math and Llemma are not. Unfortunately, we cannot fully tease apart the effect of the format of the math data (instruct vs. non-instruct) here because the format co-varies with whether code was additionally in the training mixture: all models that were additionally trained with non-instruction-formatted math data were continuously trained from models that were trained on code already. Lastly, for the experiments investigating the effect of alignment tuning, we considered models that were alignment tuned through a range of different methods and types of data, and some of the diverging findings in this experiment may be attributed to these differences. We plan to address these existing limitations through controlled training experiments in future investigations.

\subsubsection*{Acknowledgments}
We thank Abdul Rafay for running preliminary experiments as part of his master's thesis. We also thank Cookie.

\bibliography{colm2024_conference}
\bibliographystyle{colm2024_conference}

\newpage
\appendix
\section{Appendix}

\subsection{Model Details}
\label{app:model-details}
We used the \texttt{transformers} library~\citep{wolf-etal-2020-transformers} by Hugging Face for all our experiments. 
Table~\ref{tab:model_details} presents all the models along with their Hugging Face identifiers. 

For alignment-tuned models, we experimented with prompting the model with and without chat formatting (see Table~\ref{tab:in-context-prompts-chat} and Table~\ref{tab:in-context-prompts} for the different formats). Based on results over a held-out development set, we selected the best-performing prompt format. 
We find that except for Llama 2-Chat models and DeepSeek-Chat, all other alignment-tuned models performed better with the non-chat format.

\begin{table}[t]
\resizebox{1\columnwidth}{!}{
    \centering
    \setlength{\tabcolsep}{5pt}
    \begin{tabular}{lllc}
    \toprule
       
     \textbf{Model}  & \textbf{Size}  & \textbf{Hugging Face Identifier} & \textbf{Chat format} \\\midrule
      \multirow{3}{*}{Llama 2} & 7B   & \href{https://huggingface.co/meta-llama/Llama-2-7b-hf}{meta-llama/Llama-2-7b-hf} & -- \\
      & 13B   & \href{https://huggingface.co/meta-llama/Llama-2-13b-hf}{meta-llama/Llama-2-13b-hf} &  -- \\
      & 70B   & \href{https://huggingface.co/meta-llama/Llama-2-70b-hf}{meta-llama/Llama-2-70b-hf} &  --  \\ \midrule 
        \multirow{3}{*}{Llama 2-Chat} & 7B   & \href{https://huggingface.co/meta-llama/Llama-2-7b-chat-hf}{meta-llama/Llama-2-7b-chat-hf} &  \ding{51}\\
      & 13B   & \href{https://huggingface.co/meta-llama/Llama-2-13b-chat-hf}{meta-llama/Llama-2-13b-chat-hf} & \ding{51} \\
      & 70B   & \href{https://huggingface.co/meta-llama/Llama-2-70b-chat-hf}{meta-llama/Llama-2-70b-chat-hf} & \ding{51} \\\midrule

        \multirow{4}{*}{Code Llama} & 7B   & \href{https://huggingface.co/codellama/CodeLlama-7b-hf}{codellama/CodeLlama-7b-hf} & -- \\
      & 13B   & \href{https://huggingface.co/codellama/CodeLlama-13b-hf}{codellama/CodeLlama-13b-hf} & -- \\
      & 34B   & \href{https://huggingface.co/codellama/CodeLlama-34b-hf}{codellama/CodeLlama-34b-hf} & -- \\
      & 70B   & \href{https://huggingface.co/codellama/CodeLlama-70b-hf}{codellama/CodeLlama-70b-hf} & -- \\\midrule

      \multirow{4}{*}{Code Llama-Instruct} & 7B   & \href{https://huggingface.co/codellama/CodeLlama-7b-Instruct-hf}{codellama/CodeLlama-7b-Instruct-hf} & \xmark \\
      & 13B   &  \href{https://huggingface.co/codellama/CodeLlama-13b-Instruct-hf}{codellama/CodeLlama-13b-Instruct-hf} & \xmark \\
      & 34B   & \href{https://huggingface.co/codellama/CodeLlama-34b-Instruct-hf}{codellama/CodeLlama-34b-Instruct-hf} & \xmark \\
      & 70B   & \href{https://huggingface.co/codellama/CodeLlama-70b-Instruct-hf}{codellama/CodeLlama-70b-Instruct-hf} & \xmark \\\midrule
      DeepSeek & 7B & \href{https://huggingface.co/deepseek-ai/deepseek-llm-7b-base}{deepseek-ai/deepseek-llm-7b-base} & -- \\
      DeepSeek-Chat &  7B & \href{https://huggingface.co/deepseek-ai/deepseek-llm-7b-chat}{deepseek-ai/deepseek-llm-7b-chat} & \cmark \\
      DeepSeek-Coder & 7B & \href{https://huggingface.co/deepseek-ai/deepseek-coder-7b-base-v1.5}{deepseek-ai/deepseek-coder-7b-base-v1.5} & --\\
      DeepSeek-Coder-Instruct & 7B & \href{https://huggingface.co/deepseek-ai/deepseek-coder-7b-instruct-v1.5}{deepseek-ai/deepseek-coder-7b-instruct-v1.5} & \xmark \\
      DeepSeek-Math    & 7B &  \href{https://huggingface.co/deepseek-ai/deepseek-math-7b-base}{deepseek-ai/deepseek-math-7b-base} & --\\\midrule

      Gemma & 8B & \href{https://huggingface.co/google/gemma-7b}{google/gemma-7b} & -- \\
      Gemma-Instruct & 8B & \href{https://huggingface.co/google/gemma-1.1-7b-it}{google/gemma-1.1-7b-it} & \xmark \\
      CodeGemma & 8B & \href{https://huggingface.co/google/codegemma-7b}{google/codegemma-7b} & -- \\
      CodeGemma-Instruct & 8B & \href{https://huggingface.co/google/codegemma-1.1-7b-it}{google/codegemma-1.1-7b-it} & \xmark \\\midrule
    \multirow{2}{*}{Llemma} & 7B   & \href{https://huggingface.co/EleutherAI/llemma_7b}{EleutherAI/llemma\_7b} & --\\
     & 34B   & \href{https://huggingface.co/EleutherAI/llemma_34b}{EleutherAI/llemma\_34b} & -- \\\midrule

      Llama & 7B & \href{https://huggingface.co/huggyllama/llama-7b}{huggyllama/llama-7b} & -- \\
      FLoat & 7B & \href{https://huggingface.co/nikhil07prakash/float-7b}{nikhil07prakash/float-7b} & --\\
      Mistral & 7B &  \href{https://huggingface.co/mistralai/Mistral-7B-v0.1}{mistralai/Mistral-7B-v0.1} & -- \\
      OpenMathMistral & 7B & \href{https://huggingface.co/nvidia/OpenMath-Mistral-7B-v0.1}{nvidia/OpenMath-Mistral-7B-v0.1} & -- \\
      \bottomrule 
    \end{tabular}
    }
    \caption{Details of all the models evaluated in the paper. The rightmost column indicates whether the chat format was used for prompting the model. }
    \label{tab:model_details}
\end{table}

\subsection{Constrained Decoding}

\begin{figure}[!ht]
\centering
\fbox{\parbox{0.9\linewidth}{%
 \texttt{Statement: Box 0 contains( [a-zA-Z]+)*, Box 1 contains( [a-zA-Z]+)*, Box 2 contains( [a-zA-Z]+)*, Box 3 contains( [a-zA-Z]+)*, Box 4 contains( [a-zA-Z]+)*, Box 5 contains( [a-zA-Z]+)*, Box 6 contains( [a-zA-Z]+)*.%
}}}

\caption{Regular expression used for constrained decoding of entity states.}
\label{fig:regex}
\end{figure}

In our experiments, we found that the models struggled to adhere to the output format specified via the few-shot prompt examples. Luckily, the expected output can be described precisely by the regular expression shown in Figure~\ref{fig:regex}. We used the \texttt{outlines} library~\citep{Willard2023EfficientGG} which supports regex-based constrained decoding. We found a significant improvement with constrained decoding. For e.g., the performance of the Llama 2 70B model went up from 54.95 to 62.13 with constrained decoding.

\begin{table}[t]
\begin{tabular}{p{0.9\linewidth}}
    \toprule
    \textbf{2-shot prompt} \\
    \midrule

<s>[INST] <<SYS>>
Given the description after "Description:", write a true statement about all boxes and their contents to the description after "Statement:".
<</SYS>> \\\\

Description: Box 0 contains the car, Box 1 contains the cross, Box 2 contains the bag and the machine, Box 3 contains the paper and the string, Box 4 contains the bill, Box 5 contains the apple and the cash and the glass, Box 6 contains the bottle and the map. [/INST] Statement: Box 0 contains the car, Box 1 contains the cross, Box 2 contains the bag and the machine, Box 3 contains the paper and the string, Box 4 contains the bill, Box 5 contains the apple and the cash and the glass, Box 6 contains the bottle and the map. </s>\\\\

<s>[INST] Description: Box 0 contains the car, Box 1 contains the cross, Box 2 contains the bag and the machine, Box 3 contains the paper and the string, Box 4 contains the bill, Box 5 contains the apple and the cash and the glass, Box 6 contains the bottle and the map. Remove the car from Box 0. Remove the paper and the string from Box 3. Put the plane into Box 0. Move the map from Box 6 to Box 2. Remove the bill from Box 4. Put the coat into Box 3. [/INST] Statement: Box 0 contains the plane, Box 1 contains the cross, Box 2 contains the bag and the machine and the map, Box 3 contains the coat, Box 4 contains nothing, Box 5 contains the apple and the cash and the glass, Box 6 contains the bottle. </s>\\\\

<s>[INST] Description: \texttt{description}\\

    \bottomrule
\end{tabular}
\caption{Chat-formatted prompt with 2-shot in-context demonstrations.}
\label{tab:in-context-prompts-chat}
\end{table}

\subsection{Detailed Results}
\label{app:det_results}

Table~\ref{tab:det_results} presents the detailed results of all the models evaluated in this work. The results are categorized by the number of operations affecting the entity of interest. 

\begin{table}[t]
\resizebox{1\columnwidth}{!}{
    \centering
    \begin{tabular}{lccccccccc}
\toprule
\multirow{3}{*}{\shortstack[l]{\textbf{Model}}} & \multicolumn{8}{c}{\textbf{Performance split by number of operations}}\\ 
\cmidrule{2-10}
 & Overall & 0 & 1 & 2 & 3 & 4 & 5 & 6 & 7 \\
 & (5012)  & (1303) & (1410) & (1083) & (651) & (288) & (142) & (106) & (29) \\        
\midrule
Random &  21.08 & 41.06 & 17.85 & 12.70 & 11.87 & 10.58 & \phantom{1}9.16 & \phantom{1}7.51 & 12.59 \\ \midrule
Llama 2-7B &  21.31 & 65.54 & \phantom{1}4.96 & \phantom{1}7.94 & \phantom{1}4.30 & \phantom{1}5.56 & \phantom{1}4.93 & \phantom{1}5.66 & \phantom{1}3.45\\
Llama 2-7B Chat & 41.28 & 87.95 & 35.53 & 24.38 & 14.29 & 12.15 & 14.79 & \phantom{1}7.55 & \phantom{1}3.45\\
Llama 2-13B & 33.28 &  77.05 & 25.18 & 15.97 & 12.75 & \phantom{1}8.68 & \phantom{1}8.45 & 10.38 & 17.24\\
Llama 2-13B Chat & 38.05 & 77.51 & 36.67 & 19.94 & 16.28 & \phantom{1}9.03 & 11.27 & 11.32 & 13.79\\
Llama-2 70B & 62.13 & 99.00 & 71.91 & 43.67 & 31.18 & 21.88 & 25.35 & 25.47 & 27.59\\
Llama-2 70B Chat & 63.43 & 93.25 & 75.67 & 47.00 & 37.33 & 26.39 & 22.54 & 24.53 & 37.93\\
\midrule
Code Llama 7B & 31.94 & 90.87 & 13.69 & 12.28 & \phantom{1}7.99 & \phantom{1}6.60 & \phantom{1}4.93 & \phantom{1}8.49 & 13.79\\
Code Llama 7B Instruct &  41.34 & 96.16 & 36.03 & 18.19 & 12.29 & \phantom{1}6.60 & \phantom{1}5.63 & \phantom{1}3.77 & 10.34\\
Code Llama 13B & 54.79 & 95.70 & 60.78 & 32.87 & 25.35 & 20.14 & 21.83 & 26.42 & 13.79\\
Code Llama 13B Instruct & 58.14 & 97.77 & 71.13 & 35.83 & 23.35 & 17.01 & 14.79 & 19.81 & 20.69\\
Code Llama 34B & 58.26 & 94.70 & 71.28 & 33.89 & 31.18 & 21.53 & 14.79 & 19.81 & 24.14\\
Code Llama 34B Instruct & 61.47 & 95.09 & 77.09 & 39.98 & 31.03 & 22.57 & 19.01 & 20.75 & 20.69\\
Code Llama 70B & 69.77 & 99.39 & 76.60 & 54.20 & 47.31 & 40.97 & 36.62 & 43.40 & 37.93\\
Code Llama 70B Instruct & \textbf{73.66} & 98.70 & 83.40 & 59.65 & 51.92 & 46.18 & 40.85 & 39.62 & 44.83\\
\midrule
Llemma 7B & 34.12 & 84.65 & 26.81 & 13.20 & \phantom{1}8.29 & \phantom{1}4.51 & \phantom{1}6.34 & \phantom{1}9.43 & \phantom{1}0.00\\
Llemma 34B & 60.18 & 95.24 & 75.60 & 36.10 & 31.64 & 21.53 & 14.79 & 21.70 & 20.69\\
\midrule
DeepSeek 7B & 23.46 & 71.83 & \phantom{1}8.44 & \phantom{1}7.29 & \phantom{1}4.61 & \phantom{1}1.04 & \phantom{1}2.11 & \phantom{1}3.77 & \phantom{1}6.90\\
DeepSeek 7B Chat & 32.24 & 70.30 & 27.30 & 17.27 & 11.21 & 11.81 & \phantom{1}9.15 & \phantom{1}6.60 & \phantom{1}3.45\\
DeepSeek 7B Coder Base & 53.67 & 97.16 & 57.52 & 33.61 & 23.96 & 19.10 & 17.61 & \phantom{1}8.49 & 13.79\\
DeepSeek 7B Coder Instruct  & 48.76 & 91.71 & 54.54 & 26.41 & 18.13 & 13.89 & 11.27 & 17.92 & \phantom{1}3.45\\
DeepSeek 7B Math Base & 56.40 & 98.93 & 60.28 & 35.73 & 26.73 & 24.65 & 16.20 & 23.58 & 27.59\\
\midrule
Gemma-7B & 42.66 & 95.78 & 45.82 & 14.04 & \phantom{1}8.60 & \phantom{1}5.56 & \phantom{1}7.75 & \phantom{1}7.55 & \phantom{1}3.45\\
Gemma-7B Instruct & 47.79 & 93.25 & 55.96 & 21.98 & 15.51 & \phantom{1}7.99 & \phantom{1}8.45 & 11.32 & 17.24\\
CodeGemma-7B & 46.09 & 91.33 & 53.76 & 21.70 & 13.67 & \phantom{1}6.94 & \phantom{1}5.63 & \phantom{1}8.49 & \phantom{1}3.45\\
CodeGemma-7B Instruct & 53.49 & 93.48 & 69.57 & 26.22 & 19.82 & 12.50 & 12.68 & 11.32 & 10.34\\
\midrule
Mistral 7B & 45.67 & 96.01 & 50.28 & 18.56 & 11.37 & \phantom{1}8.68 & 12.68 & \phantom{1}7.55 & 10.34\\
OpenMathMistral & 38.25 & 90.94 & 34.40 & 14.22 & \phantom{1}8.76 & \phantom{1}5.90 & \phantom{1}4.93 & \phantom{1}5.66 & 20.69\\
\midrule
LLama 7B & 28.67 & 97.93 & \phantom{1}0.71 & 10.53 & \phantom{1}2.92 & \phantom{1}3.82 & \phantom{1}3.52 & \phantom{1}0.94 & \phantom{1}3.45\\
Float 7B Instruct & 27.55 & 89.33 & \phantom{1}1.56 & 12.37 & \phantom{1}5.07 & \phantom{1}5.21 & \phantom{1}2.82 & \phantom{1}4.72 & 13.79\\
\bottomrule
\end{tabular}
}
    \caption{Entity Tracking performance of models categorized by the number of operations affecting the entity of interest. The count of test set instances with the number of operations affecting an entity is indicated in parentheses below the corresponding column title.}
    \label{tab:det_results}
\end{table}

\end{document}